\documentclass{article}

\usepackage[final, nonatbib]{neurips_2024}

\usepackage[utf8]{inputenc} 
\usepackage[T1]{fontenc}    
\usepackage[bookmarks=false]{hyperref}
\usepackage{url}            
\usepackage{booktabs}       
\usepackage{amsfonts}       
\usepackage{nicefrac}       
\usepackage{microtype}      
\usepackage{xcolor}         
\usepackage{graphicx}
\usepackage{amsmath}
\usepackage{multirow}

\title{Boosting Unsupervised Segmentation Learning}

\author{%
  Alp Eren ~Sari\\
  University of Bern\\
  Switzerland\\
  \texttt{alp.sari@unibe.ch} \\
  \And
  Francesco ~Locatello \\
  Institute of Science and Technology Austria\\ 
  Austria \\
  \texttt{francesco.locatello@ista.ac.at} \\
  \AND
  Paolo ~Favaro \\
  University of Bern\\
  Switzerland\\
  \texttt{paolo.favaro@unibe.ch} \\
}

\begin{document}

\maketitle

\begin{abstract}
  We present two practical improvement techniques for unsupervised segmentation learning. These techniques address limitations in the resolution and accuracy of predicted segmentation maps of recent state-of-the-art methods. Firstly, we leverage image post-processing techniques such as guided filtering to refine the output masks, improving accuracy while avoiding substantial computational costs. Secondly, we introduce a multi-scale consistency criterion, based on a teacher-student training scheme. This criterion matches segmentation masks predicted from regions of the input image extracted at different resolutions to each other. Experimental results on several benchmarks used in unsupervised segmentation learning demonstrate the effectiveness of our proposed techniques.
\end{abstract}

\section{Introduction}
\label{sec:intro}

The task of segmenting objects in images is a fundamental first step in extracting semantic information. Current approaches \cite{kirillov2023segment, ke2024segment} have reached remarkable performance and capabilities by combining various annotation modalities (text, points, bounding boxes, and segmentation masks) and by scaling the training to large datasets. However, recently, several methods \cite{hamilton2021unsupervised, zadaianchuk2022unsupervised, seitzer2023bridging, ravindran2023sempart,bielski2022move,wang2023tokencut} have shown that it is possible to learn to segment objects in an unsupervised manner, i.e., without the guidance of any human annotation. Besides the intrinsic scientific significance of such unsupervised learning methods, their main advantage is that they could potentially be scaled to extremely large datasets and across multiple imaging modalities with limited human effort.

Despite encouraging results, these methods seem to be either limited in resolution \cite{ravindran2023sempart} or to rely on difficult training schemes \cite{bielski2022move}.
Since many of the existing methods \cite{melas2022deep,wang2023tokencut,ravindran2023sempart} rely on DINO features \cite{caron2021emerging}, which reduce the initial image resolution by a factor of $8$ or $16$, we introduce two rather general ``tricks'' to enhance the resolution of predicted segmentation maps and demonstrate their capabilities on the recent state of the art method \emph{Sempart} for unsupervised segmentation learning \cite{ravindran2023sempart}.
\footnote{\url{https://github.com/alpErenSari/segmentation-tricks}}
Our first trick is to use post-processing to enhance the resolution of the output masks. We show that well-established image processing methods such as guided filtering \cite{he2012guided} can easily improve the accuracy of the predicted segmentation masks when the input image luminance is used as the filter guidance. 

Our second trick is a general criterion to enhance the resolution of dense predictions. We introduce a \emph{multi-scale consistency} criterion on the predicted segmentation mask. We do so by using a teacher-student training scheme, where the teacher network takes a zoomed-in region of an image as input and the student network takes the whole image as input. Then, we zoom in on the same region of the predicted segmentation mask from the student network's prediction and match it to the prediction from the teacher. Because we use the output of the teacher network as a target, we only backpropagate through the student network. 

In summary, our contributions are
\begin{itemize}
    \item We achieve new state-of-the-art (SotA) results in unsupervised saliency segmentation on the DUT-OMRON \cite{yang2013saliency}, DUTS-TE \cite{wang2017learning}, and ECSSD \cite{shi2015hierarchical} datasets;
    \item We introduce two novel and general techniques to enhance the resolution of the segmentation masks predicted by existing SotA that are computationally efficient;
    \item Our techniques are easy to apply and have been thoroughly tested on several benchmarks and in combination with different methods; the code will be made available upon publication.
\end{itemize}


\section{Related Work}
\label{sec:related}
\noindent\textbf{Spectral Methods.} Graph representations are commonly utilized in computer vision and computer graphics for partitioning problems. \cite{shi2000normalized} propose the \textit{normalized cuts} method for image segmentation to handle the bias towards small-isolated segments in the minimum graph cut method \cite{wu1993optimal}.\\ 
\noindent\textbf{Self-supervised representations (SSL).} The  idea of training deep networks without any supervision to obtain effective image representations gained popularity with the success of  \cite{chen2020simple, he2020momentum, caron2020unsupervised, caron2021emerging, he2022masked}. In particular, DINO \cite{caron2021emerging} features based on ViT \cite{dosovitskiy2020image} demonstrate capabilities beyond image classification. \\
\noindent\textbf{Unsupervised Semantic Segmentation.} Several recent approaches, such as \cite{simeoni2021localizing, melas2022deep, wang2023tokencut}, explore the use of DINO features and graph cuts for semantically meaningful image segmentation. However, these methods often exhibit high computational complexity and relatively low segmentation accuracy. In contrast, \cite{bielski2019emergence, bielski2022move} rely on the local ``shiftability'' of segmented objects, but face challenges due to the complexity of their adversarial training schemes. \cite{melas2021finding} explores pre-trained generative adversarial networks (GAN). However, it is also challenging to generalize this approach due to the limitations of existing GANs. Other methods, like \cite{van2021unsupervised, simeoni2023unsupervised, shin2022unsupervised, ravindran2023sempart}, leverage various strategies, such as push-pull forces between pixels, and pseudo-masks, to distill segmentation networks. However, these approaches often suffer from low-resolution image features, limiting the quality of the generated masks.



\section{Two Tricks}
\label{sec:tricks}
\begin{figure}[t]
  \centering
  \includegraphics[width=0.8\linewidth]{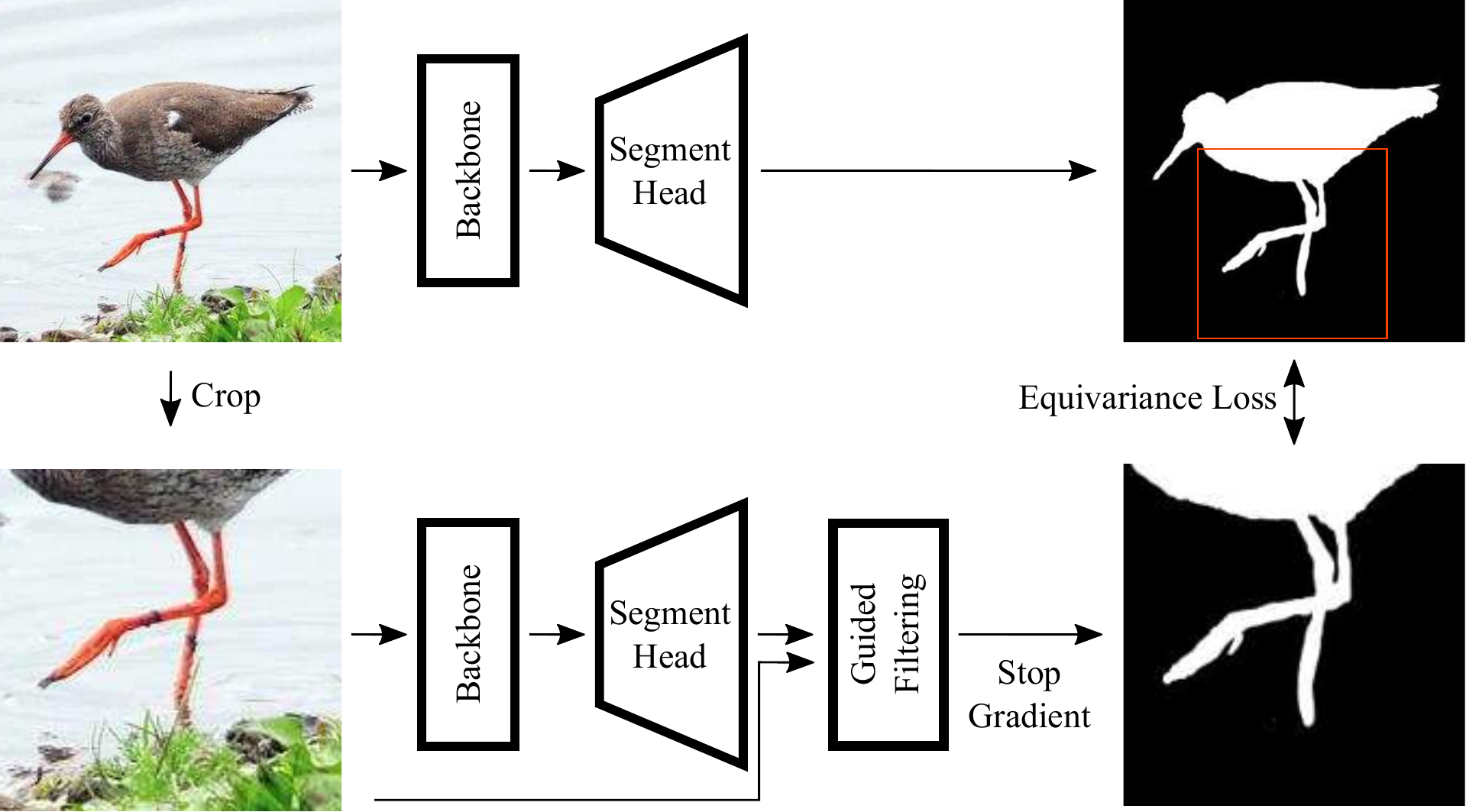}
  \caption{The illustration of the proposed multi-scale consistency procedure. First, we predict a segmentation from an input image. Then, we randomly crop a portion of the input image, predict a more detailed segmentation mask, refine this prediction with guided filtering, and apply a stop-gradient operation to the final mask to prevent a mask prediction collapse. Finally, we calculate the mean squared error between the corresponding region of the initial mask and the detailed target mask.}
  \label{fig:main_crop}
\end{figure}

We introduce two tricks that can be incorporated into different unsupervised segmentation methods in a flexible manner: guided filtering\cite{he2012guided} and random cropping-based equivariance loss. \\
\noindent\textbf{Guided Filtering.} First, the RGB image -whose saliency mask is predicted- is converted to grayscale. Afterward, the guided filter \cite{he2012guided} is applied to the predicted segmentation mask using the grayscale image as guidance. \\
\noindent\textbf{Multi-Scale Consistency.} One challenge associated with incorporating a segmentation head over a ViT  \cite{dosovitskiy2020image} feature map (as with DINO) is the loss of details, particularly for small objects or structures within the image. Sempart \cite{ravindran2023sempart} addresses this issue by feeding the input image into the segmentation head and by applying a total variation loss on the resulting high-resolution mask. We aim to tackle this problem by introducing an equivariance loss based on random cropping depicted in Fig. \ref{fig:main_crop}. 
The proposed equivariance loss is computed through the following steps:
\begin{enumerate}
    \item Given an input image and a segmentation framework denoted as $I$ and $f(\cdot)$ respectively, predict a mask denoted as $\hat{s}$. This process can be expressed as $f(I) = \hat{s}$.
    \item Randomly crop the same region of the input image and the predicted mask and enlarge both of them to the original input image size. This operation on $I$, $\hat{s}$ can be expressed as $\text{CU}(I, \hat{s}) = \{I_c, s_c\}$ where $\text{CU}(\cdot, \cdot)$, $I_c$, and $s_c$ are the crop-zoom function, the resultant image and mask.
    \item Use the resultant image to predict a mask and refine this mask with guided image filtering, which can be expressed as $f(I_c) = \hat{s}_c$.
    \item Apply stop gradient afterward to prevent a collapse of the predicted masks, which we denote as $\text{SG}[\text{GF}(\hat{s}_c)] = s_{target}$, where $s_{target}$ is the final target for our loss, and $\text{SG}[\cdot]$ is the stop gradient operation.
    \item Calculate the mean squared error between the new prediction and the corresponding region of the first segmentation prediction as $\mathcal{L}_{eq} = \frac{1}{hw}{\lVert s_c - s_{target} \rVert}^2_2$ where $h$ and $w$ are the height and width of the corresponding masks $\hat{s}_c$ and $s_{target}$.
\end{enumerate}
In our experiments, we added $\mathcal{L}_{eq}$ multiplied with the parameter $\lambda_{eq}$ to the overall loss of Sempart \cite{ravindran2023sempart}. We also used the cropped image to train the network.

\section{Experimental Results}
\label{sec:results}
\begin{table}[t]
\caption{The unsupervised saliency segmentation results on DUT-OMRON \cite{yang2013saliency}, DUTS-TE \cite{wang2017learning}, and ECSSD \cite{shi2015hierarchical}. The best results are indicated in \textbf{bold}. Sempart$^\ast$ indicates our implementation of Sempart \cite{ravindran2023sempart} since the code is currently unavailable. ``Ours'' indicates the application of the proposed trick over the Sempart baseline.  We demonstrate improvements over the baseline implementation we have.}
 \centering
 \resizebox{\columnwidth}{!}{%
  \begin{tabular}{l|c|c|c|c|c|c|c|c|c}
    \toprule
    \multirow{3}{*}{Model \hspace{1mm}} &
      \multicolumn{3}{c|}{DUT-OMRON \cite{yang2013saliency} \hspace{1mm}} &
      \multicolumn{3}{c|}{DUTS-TE \cite{wang2017learning} \hspace{1mm}} &
      \multicolumn{3}{c}{ECSSD \cite{shi2015hierarchical} \hspace{1mm}} \\ \cline{2-10}
      & { Acc } & {IoU} & {$maxF_{\beta}$} & { Acc } & {IoU} & {$maxF_{\beta}$} & {Acc } & {IoU} & {$maxF_{\beta}$} \\
      \midrule
    LOST \cite{simeoni2021localizing} & .797 & .410 & .473 & .871 & .518 & .611 & .895 & .654 & .758 \\
    TokenCut \cite{wang2023tokencut} & .880 & .533 & .600 & .903 & .576 & .672 & .918 & .712 & .803 \\
    FOUND \cite{simeoni2023unsupervised} - single & .920 & .586 & .683 & .939 & .637 & .733 & .912 & .793 & .946 \\
    MOVE \cite{bielski2022move} & .923 & .615 & .712 & .950 & .713 & .815 & .954 & .830 & .916 \\
    Sempart \cite{ravindran2023sempart} & \textbf{.932} & .668 & .764 & \textbf{.959} & .749 & .867 & \textbf{.964} & .855 & \textbf{.947} \\
    Sempart$^\ast$  & .908 & .664 & .758 & .949 & .761 & .856 & .950 & .853 & .925 \\
    Ours & .920 & \textbf{.684} & \textbf{.765} & .956 & \textbf{.787} & \textbf{.869} & .952 & \textbf{.862} & .931 \\
    \midrule
    MOVE \cite{bielski2022move} + BF & .931 & .636 & .734 & .951 & .687 & .821 & .953 & .801 & .916 \\
    Sempart + BF & \textbf{.942} & \textbf{.698} & \textbf{.799} & \textbf{.958} & \textbf{.749} & \textbf{.879} & \textbf{.963} & \textbf{.850} & \textbf{.944} \\
    Sempart$^\ast$ + BF & .914 & .689 & .770 & .950 & .765 & .857 & .950 & .849 & .925 \\
    Ours + BF & .920 & .675 & .763 & .950 & .742 & .855 & .945 & .830 & .920 \\
    \midrule
    \textsc{SelfMask} w/ MOVE & .933 & .666 & .756 & .954 & .728 & .829 & .956 & .835 & .921 \\
    \textsc{SelfMask} w/ Sempart & \textbf{.942} & \textbf{.698} & \textbf{.799} & \textbf{.958} & \textbf{.749} & .879 & \textbf{.963} & .850 & .944 \\
    \textsc{SelfMask} w/ Sempart$^\ast$ & .923 & .674 & .768 & .953 & .743 & .866 & .962 & \textbf{.854} & .942 \\
    \textsc{SelfMask} w/ Ours & .935 & .685 & .790 & .956 & .747 & \textbf{.881} & .962 & .852 & \textbf{.948} \\
    \bottomrule
  \end{tabular}%
}
  \label{table:main_results}
\end{table}
We evaluate the tricks we propose on Sempart \cite{ravindran2023sempart} since it is the current state-of-the-art unsupervised segmentation learning method, it trains stably, and it converges quickly. The evaluations are performed for saliency segmentation and single object detection since Sempart \cite{ravindran2023sempart} and previous works were evaluated on these tasks. All the evaluations were performed using the fine branch of Sempart \cite{ravindran2023sempart}. The improved baseline with the proposed tricks is denoted as ``ours''.
\subsection{Unsupervised Saliency Segmentation}
\noindent\textbf{Datasets.} We used the DUTS-TR \cite{wang2017learning} for training in every experiment, as \cite{shin2022unsupervised, bielski2022move, ravindran2023sempart}. To evaluate the saliency segmentation, we used DUT-OMRON \cite{yang2013saliency}, DUTS-TE \cite{wang2017learning}, and ECSSD \cite{shi2015hierarchical} datasets. \\
\noindent\textbf{Evaluation.} Three metrics are reported in our work as \cite{shin2022unsupervised, bielski2022move, ravindran2023sempart}, which are pixel mask accuracy (Acc), intersection over union (IoU), and $\max \textit{F}_{\beta}$ \cite{wang2023tokencut}. \\
\noindent\textbf{Results.} In Table \ref{table:main_results}, we present a comprehensive comparison of our results with the recent state-of-the-art method Sempart \cite{ravindran2023sempart}. Our implementation, denoted as ``Sempart$^\ast$'', undergoes evaluation across three distinct criteria: baseline results, post-processed baseline results employing bilateral filtering \cite{barron2016fast}, and outcomes derived from \textsc{SelfMask} \cite{shin2022unsupervised}, trained with pseudo-masks generated by various methods. The comparison reveals the efficacy of our approach in advancing unsupervised saliency segmentation. 
\par Our results show state-of-the-art Intersection over Union (IoU) performance across all three datasets. When considering the notable gaps between our implemented baseline and the reported baseline, particularly in terms of accuracy (Acc) and $\max \textit{F}_{\beta}$ metrics, it becomes evident that our approach has the potential to establish a new state-of-the-art benchmark with precisely the same baseline.

\subsection{Single Object Detection}
\noindent\textbf{Datasets.} The evaluation of single object detection was done on three datasets, which are the train split of COCO20K \cite{lin2014microsoft}, the training and validation splits of Pascal VOC07 \cite{everingham2010pascal}, and Pascal VOC12 \cite{everingham2012pascal}. \\
\noindent\textbf{Evaluation.} We follow the evaluation protocol of the previous methods \cite{wang2023tokencut, bielski2022move, ravindran2023sempart} and report \textit{Correct Localization} (CorLoc) metric. \\
\noindent\textbf{Results.} In Table \ref{table:main_detection}, we conduct a comprehensive comparison between our results and the recent state-of-the-art MOVE \cite{bielski2022move}, as well as our baseline in object detection. Our tricks significantly enhance the baseline across all three datasets considered. This improvement is evident in detection accuracy, underscoring the effectiveness of our approach in advancing unsupervised object detection.

\begin{table}[t]
\caption{Comparison of single object detection on Pascal VOC07 \cite{everingham2010pascal}, Pascal VOC12 \cite{everingham2012pascal}, COCO20K \cite{lin2014microsoft} with CorLoc metric. The best results are indicated in \textbf{bold}. Sempart$^\ast$ indicates our implementation of Sempart \cite{ravindran2023sempart}. ($\uparrow z$) indicate the improvements over the baseline.}
 \centering
  \begin{tabular}{l|c|c|c}
    \toprule
    Model & \hspace{5mm} VOC07 \cite{everingham2010pascal} \hspace{5mm} &  \hspace{5mm} VOC12 \cite{everingham2012pascal}  \hspace{5mm} & COCO20K \cite{lin2014microsoft} \\
      \midrule
    LOST \cite{simeoni2021localizing} & 61.9 & 64.0 & 50.7 \\
    Deep Spectral \cite{melas2022deep} & 62.7 & 66.4 & 50.7 \\
    TokenCut \cite{wang2023tokencut} & 68.8 & 72.1 & 58.8 \\
    MOVE \cite{bielski2022move} & \textbf{76.0} & \textbf{78.8} & \textbf{66.6} \\
    Sempart \cite{ravindran2023sempart} & 75.1 & 76.8 & 66.4 \\
    Sempart$^\ast$ & 71.7 & 75.9 & 61.3 \\
    Ours & \hspace{9mm} 74.4 ($\uparrow 2.7$) &  \hspace{9mm} 77.0  ($\uparrow 1.1$) &  \hspace{9mm} 63.4  ($\uparrow 2.1$) \\
    \bottomrule
  \end{tabular}
  \label{table:main_detection}
\end{table}

\section{Conclusion}
\label{sec:conclusion}
We introduce two practical methods that can be flexibly incorporated into unsupervised image segmentation. We demonstrate their effectiveness with state-of-the-art results on unsupervised saliency segmentation tasks, combined with the recent state-of-the-art baseline. Moreover, in the appendix, we demonstrate that these two methods can be effective even if the baseline was trained with different backbones and combined with other recent unsupervised segmentation methods. We also include related background information, more qualitative results illustrating the gains of our methods, and an ablation study in unsupervised saliency segmentation and unsupervised object discovery in the appendix.


\bibliographystyle{plain}
\bibliography{main}


\appendix
\section{Additional Background}
\label{sec:additional_background}
In this section we briefly revise some basic definitions to make the paper self-contained and also to provide a context for our proposed techniques.

\noindent\textbf{Normalized Cuts.} Normalized cuts \cite{shi2000normalized} is a method to partition a weighted undirected graph $G = (V, E)$ with weights $w_{ij}$ representing the similarity between nodes $v_i$ and $v_j$, into partitions $A$ and $B$ by minimizing the following loss\\
\begin{align}
    \text{Ncut}(A, B) &= \frac{\text{cut}(A, B)}{\text{assoc}(A, V)} + \frac{\text{cut}(B, A)}{\text{assoc}(B, V)},\\
    \text{where }     \text{cut}(A, B) &\doteq\!\!\! \sum\limits_{u \in A, v \in B}\!\!\!w_{uv} \quad \text{ and } \quad
    \text{assoc}(A, V) \doteq \!\!\!\sum\limits_{u \in A, t \in V}\!\!\!w_{ut}.\nonumber
\end{align}
$\text{cut}(B, A)$, and $\text{assoc}(B, V)$ are also defined similarly. By minimizing Ncut with respect to $A$ and $B$, the connectivity between partitions $A$ and $B$ is minimized and the connectivity within the partitions is maximized. The weights are collected in an \emph{affinity matrix} $W$ such that its $i,j$-th entry $W_{i,j} = w_{ij}$.
The minimization of normalized cuts loss is NP-complete; hence, \cite{shi2000normalized} proposed a relaxation of the proposed loss which leads to a generalized eigenvalue problem. \\
\noindent\textbf{TokenCut.} \cite{caron2021emerging} shows that the attention maps in DINO strongly relate to the location and shape of the object present in the input image. Thus, TokenCut \cite{wang2023tokencut} proposes to utilize DINO features for semantic segmentation combined with normalized cuts as a \emph{post-processing} step. 
TokenCut \cite{wang2023tokencut} minimizes the normalized cuts' loss with the following affinity matrix
\begin{align}
w_{ij} = \left\{
\begin{array}{l}
    1 \mid \langle F_{v_i},F_{v_j}\rangle > \tau \\
    \epsilon \mid \text{otherwise}.
\end{array}\label{ncutaff}
\right.
\end{align}
where $F_{v_i}$ is the DINO feature corresponding to the node $v_i$, and $\tau$ and $\epsilon$ are scalar thresholds to be tuned. \\
\noindent\textbf{Sempart.} Sempart \cite{ravindran2023sempart} instead leverages DINO features by training a convolutional neural networks (CNN) segmentation head to minimize a loss based on the normalized cut cost. The segmentation head generates two segmentation predictions: one,  $S_\text{coarse} \in [0, 1]^{|V|}$, at a low-resolution and one, $S_\text{fine} \in [0, 1]^{HW}$, at high-resolution, where $|V|$ is the number of DINO features, and $H$ and $W$ are the height and width of the input image. $S_\text{coarse}$ minimizes the following variation of the normalized cuts' loss
\begin{align}
    \label{eq:sempart_ncut}
    \mathcal{L}_{\text{Ncut}} = \text{Ncut}(A, B) = \frac{S^{T}W(1 - S)}{S^T W\mathbf{1}} +  \frac{(1-S)^{T}WS}{(1 - S)^T W\mathbf{1}},
\end{align}
where $\mathbf{1}$ is a vector of ones.
Eq. \ref{eq:sempart_ncut} is optimized only for the low-resolution mask $S_{\text{coarse}}$. High-resolution mask $S_\text{fine}$ is guided through a consistency loss 
\begin{align}
\mathcal{L}_{SR} = {\lVert \text{down}(S_\text{fine}) - S_\text{coarse} \rVert}^2_2,    
\end{align}
which matches the downsampled $S_\text{fine}$ and $S_\text{coarse}$, and $\mathcal{L}_\text{GTV-fine}$ is a graph total variation (GTV) loss that works on the high-resolution mask. An additional GTV loss $\mathcal{L}_\text{GTV-coarse}$ computed with $S_\text{coarse}$ is also used for additional guidance. $\mathcal{L}_\text{GTV-fine}$ and $\mathcal{L}_\text{GTV-coarse}$ are given as
\begin{align}
    \mathcal{L}_\text{GTV-fine/coarse} &= \frac{1}{2}\sum\limits_{(i,j)\in E} a_{ij} (s_i - s_j)^2, \\
    a_{ij}^\text{fine} &= \exp(-{\lVert x_i - x_j \rVert}^2_2 / \sigma), \\
    a_{ij}^\text{coarse} &= w_{ij}\mathbf{I}\{ i \in \mathcal{N}(j) \},
\end{align}
where $\mathcal{N}(j)$ is set of nodes adjacent to node j, and $\mathbf{I}$ is the indicator function. The only difference between $\mathcal{L}_\text{GTV-fine}$ and $\mathcal{L}_\text{GTV-coarse}$ is that $x_i$ is a pixel value from the input image for fine GTV, but it is the DINO feature for the coarse GTV. Finally, the segmentation head is trained with the combined loss
\begin{align}
\label{eq:loss}
    \mathcal{L}_\text{sempart} = \mathcal{L}_{\text{Ncut}} + \lambda_\text{GTV-coarse}\mathcal{L}_\text{GTV-coarse} +  
    \lambda_\text{GTV-fine}\mathcal{L}_\text{GTV-fine} + \lambda_\text{SR}\mathcal{L}_\text{SR}.
\end{align}

\section{Additional Results}
\label{sec:additional_results}
\subsection{Unsupervised Saliency Segmentation}
\begin{figure}[t]
  \centering
  \includegraphics[width=\linewidth]{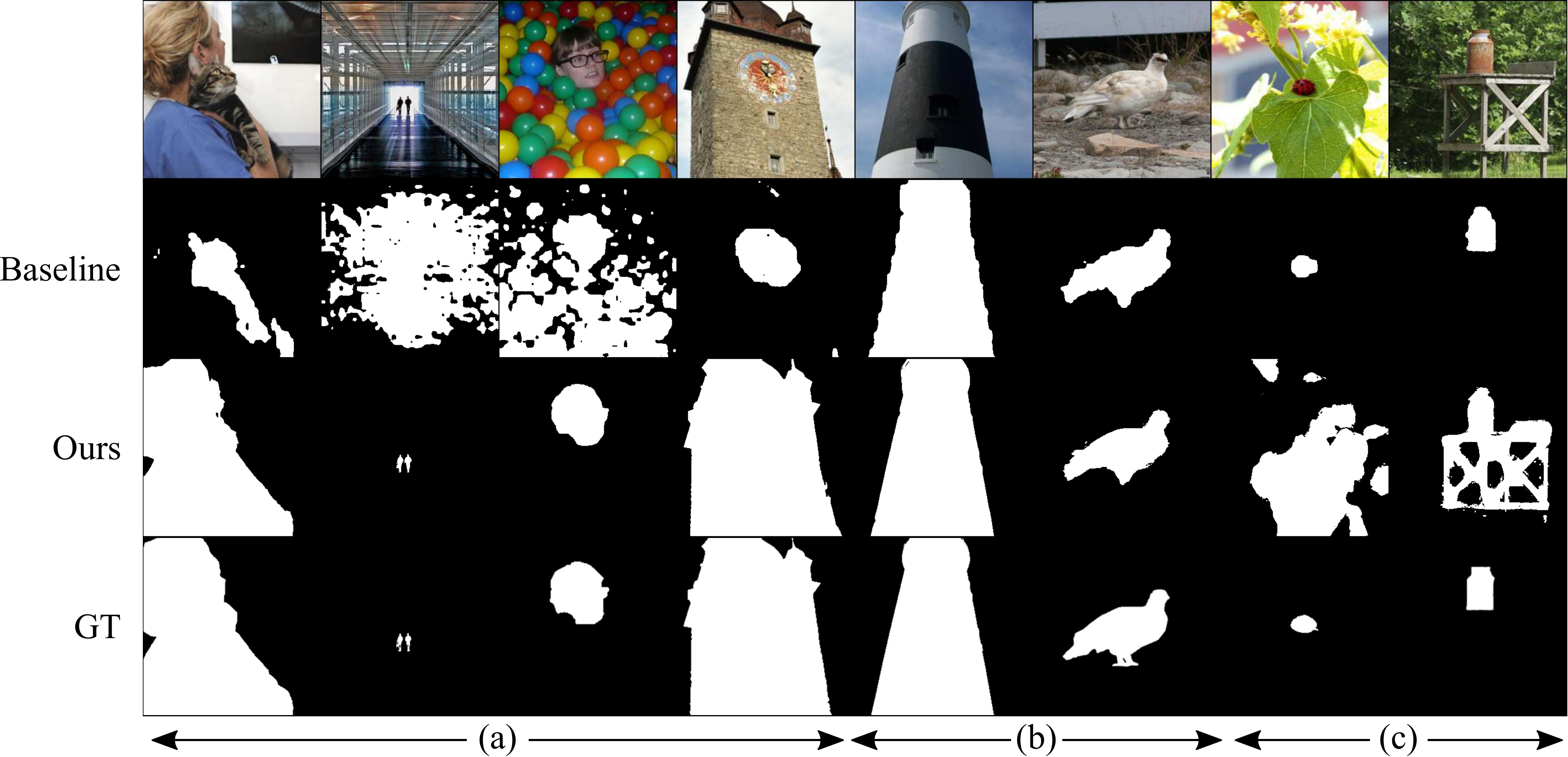}
  \caption{A qualitative comparison between the baseline and the improved segmentation with our proposed tricks (Ours). The first row shows images sampled from DUTS-TE \cite{wang2017learning}, the second row shows the baseline segmentation predictions, the third row shows the segmentation results with our tricks, and the last row shows the ground truth segmentation masks. In (a) we show 4 image samples where we achieve a significant improvement over the baseline; in (b) we show two image samples where we achieve the same results as in the baseline; in (c) we show two image samples where the tricks make the segmentation masks worse (relative to the ground truth mask). We point out that some of the incorrect masks may also be due to the inherent ambiguity of the unsupervised segmentation task.}
  \label{fig:main_segmentation}
\end{figure}

We provide some qualitative results in Fig. \ref{fig:main_segmentation} for the cases where the segmentation is improved, unaffected, and degraded. 
\subsection{Single Object Detection}
\begin{figure}[t]
  \centering
  \includegraphics[width=\linewidth]{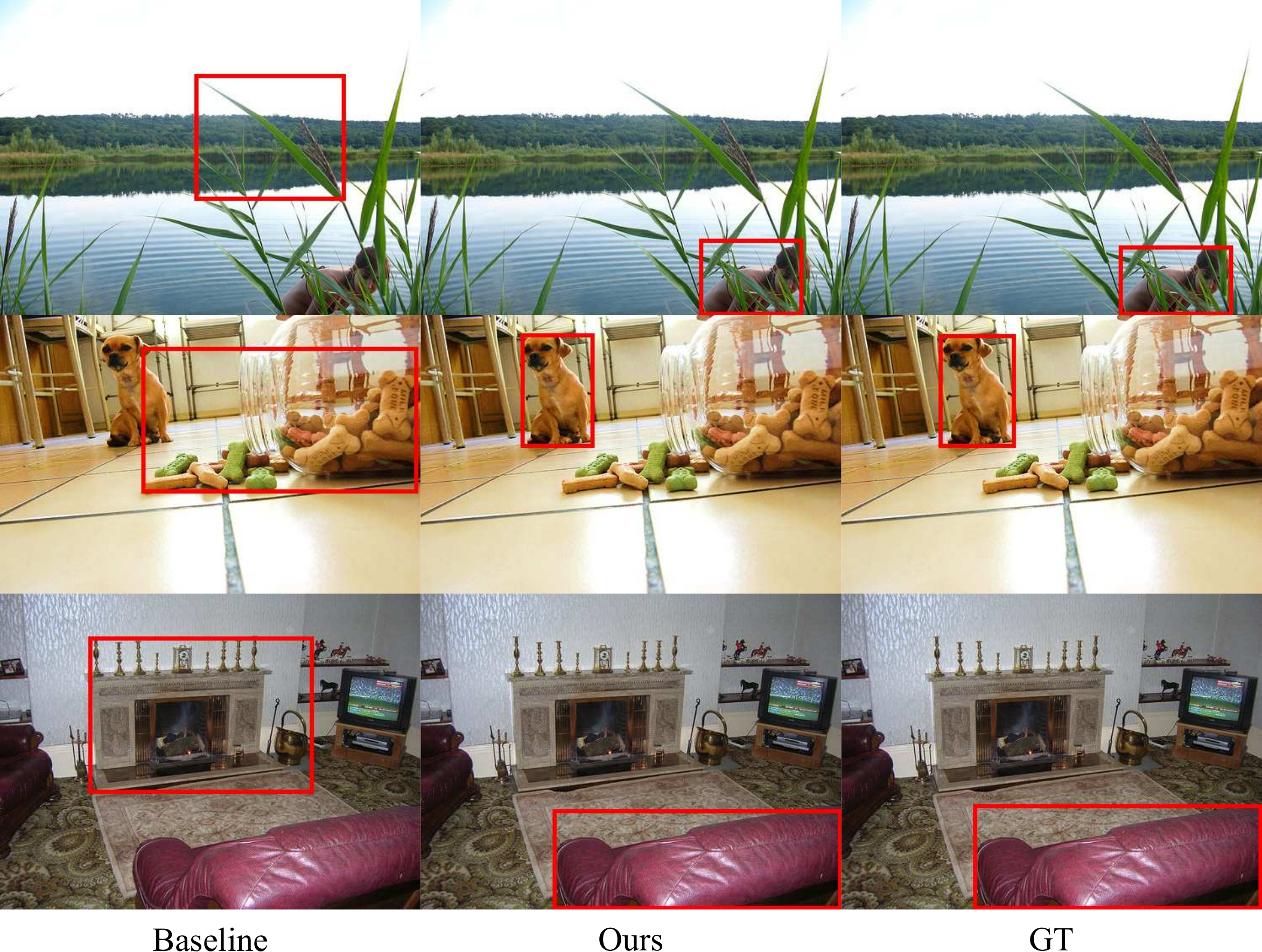}
  \caption{Some examples cases from Pascal VOC12 \cite{everingham2012pascal} where our tricks improve object detection over baseline.}
  \label{fig:main_detection}
\end{figure}

We share the implementation details of the single object detection evaluation and some qualitative results here to illustrate how our tricks boost object detection. Specifically, Fig. \ref{fig:main_detection} visually depicts how random cropping facilitates the detection of small objects and, in some instances, objects with occlusion. 
\noindent\textbf{Evaluation Details.} First, the predicted mask is split into parts if it's not fully connected. Then, we select the mask with the largest bounding box as our object prediction and calculate the IoU between our prediction and the ground-truth bounding box. Finally, the percentage of the IoUs higher than 0.5 is calculated, known as the \textit{Correct Localization} (CorLoc) metric. \\

\subsection{Ablations}
\label{results:subsection:ablations}
We conducted various ablation experiments to demonstrate the effectiveness of the proposed enhancements. These experiments involved evaluating the impact of the proposed techniques compared to the baseline in unsupervised saliency segmentation and single object detection. Additionally, we performed an ablation study on the guided filtering, comparing its performance with recent methods such as TokenCut \cite{wang2023tokencut} and MOVE \cite{bielski2022move} in unsupervised saliency segmentation. For the saliency segmentation ablation, we trained different baselines using various SSL backbones and demonstrated that our techniques consistently outperformed the baselines across different backbones. \\
\noindent\textbf{The tricks on saliency segmentation.} In Table \ref{table:ablation_segmentation}, we present the Intersection over Union (IoU) results obtained from the evaluation of unsupervised saliency segmentation on three datasets: DUT-OMRON \cite{yang2013saliency}, DUTS-TE \cite{wang2017learning}, and ECSSD \cite{shi2015hierarchical}. The original baseline is trained with DINO \cite{caron2021emerging} ViT-S/8 as the backbone. Additionally, we provide results for baselines trained with DINO ViT-S/16 and IBOT \cite{zhou2021image} ViT-S/16 as the backbones. Our findings indicate that across different backbones, the proposed techniques significantly enhance segmentation performance. Notably, guided filtering emerges as a key contributor to the improved segmentation results. \\
\begin{table}[t]
\caption{The ablations on unsupervised saliency segmentation with mIoU on three datasets; DUT-OMRON \cite{yang2013saliency}, DUTS-TE \cite{wang2017learning}, and ECSSD \cite{shi2015hierarchical}. The best results are indicated with \textbf{bold}.}
 \centering
  \begin{tabular}{c|c|c|c|c|c}
    \toprule
    Backbone & Crop & GF & DUT-OMRON \cite{yang2013saliency} & DUTS-TE \cite{wang2017learning} & ECSSD \cite{shi2015hierarchical} \\
    \midrule
     \multirow{4}{*}{\parbox{1.5cm}{DINO \cite{caron2021emerging} ViT-S/8}} & $\times$ & $\times$ & .664 & .761 & .853 \\
     & \checkmark & $\times$ & .670 & .772 & .852 \\
     & $\times$ & \checkmark & \textbf{.684} & .784 & \textbf{.869} \\
     & \checkmark & \checkmark & \textbf{.684} & \textbf{.787} & .862 \\
     \midrule
     \multirow{4}{*}{\parbox{1.5cm}{DINO \cite{caron2021emerging} ViT-S/16}} & $\times$ & $\times$ & .604 & .692 & .784 \\
     & \checkmark & $\times$ & .647 & .710 & .812 \\
     & $\times$ & \checkmark & .608 & .697 & .788 \\
     & \checkmark & \checkmark & \textbf{.652} & \textbf{.714} & \textbf{.816} \\
     \midrule
     \multirow{4}{*}{\parbox{1.5cm}{IBOT \cite{zhou2021image} ViT-S/16}} & $\times$ & $\times$ & .584 & .650 & .793 \\
     & \checkmark & $\times$ & .588 & .654 & .797 \\
     & $\times$ & \checkmark & .600 & .664 & .806 \\
     & \checkmark & \checkmark & \textbf{.604} & \textbf{.668} & \textbf{.811} \\
    \bottomrule
  \end{tabular}
  \label{table:ablation_segmentation}
\end{table}

\begin{table}[t]
\caption{The ablations on unsupervised single object detection with CorLoc on three datasets; 
Pascal VOC07 \cite{everingham2010pascal}, Pascal VOC12 \cite{everingham2012pascal}, COCO20K\cite{lin2014microsoft}. The best results are indicated with \textbf{bold}.}
 \centering
  \begin{tabular}{c|c|c|c|c}
    \toprule
    Crop & GF & VOC07 \cite{everingham2010pascal} &  VOC12 \cite{everingham2012pascal} & COCO20K \cite{lin2014microsoft} \\
      \midrule
     
     $\times$ & $\times$ & 71.7 & 75.9 & 61.3 \\
     \checkmark & $\times$ & 74.2 & \textbf{77.0} & \textbf{64.3} \\
     $\times$ & \checkmark & 71.7 & 75.7 & 61.3 \\
     \checkmark & \checkmark & \textbf{74.4} & \textbf{77.0} & 63.4 \\
    \bottomrule
  \end{tabular}
  \label{table:ablation_detection}
\end{table}

\noindent\textbf{The tricks on single object detection.} In Table \ref{table:ablation_detection}, we demonstrate the enhancement achieved by our techniques in object detection based on the CorLoc metric. The evaluation is performed on datasets comprising the train and validation splits of Pascal VOC07 \cite{everingham2010pascal} and Pascal VOC12 \cite{everingham2012pascal}, as well as the train split of COCO20K \cite{lin2014microsoft}. Notably, unlike the case of saliency segmentation, we observe that random cropping yields the most significant improvement in object detection performance. \\
\noindent\textbf{Guided filtering with different methods.} In Table \ref{table:ablation_gf}, our findings indicate that the application of guided filtering significantly enhances the performance of recent saliency segmentation methods TokenCut \cite{wang2023tokencut} and MOVE \cite{bielski2022move}. We present results Acc, IoU, and the $max \textit{F}_{\beta}$ metric across evaluation datasets including DUT-OMRON \cite{yang2013saliency}, DUTS-TE \cite{wang2017learning}, and ECSSD \cite{shi2015hierarchical}. Comparative analysis with baseline methods, incorporating bilateral filtering \cite{barron2016fast}, is also included. Notably, the results in Table \ref{table:ablation_gf} demonstrate consistent improvement in all metrics with guided filtering, which contrasts with the performance of bilateral filtering.

\begin{table}[t]
\caption{The ablations of guided filtering \cite{he2012guided} and bilateral filtering \cite{barron2016fast} with TokenCut \cite{wang2023tokencut} and MOVE \cite{bielski2022move} on unsupervised saliency segmentation with same datasets and metrics in Table \ref{table:main_results}. The best results for each method are indicated with \textbf{bold}.}
 \centering
  \begin{tabular}{l|c|c|c|c|c|c|c|c|c}
    \toprule
    \multirow{3}{*}{Model \hspace{1mm}} &
      \multicolumn{3}{c|}{DUT-OMRON \cite{yang2013saliency} \hspace{1mm}} &
      \multicolumn{3}{c|}{DUTS-TE \cite{wang2017learning} \hspace{1mm}} &
      \multicolumn{3}{c}{ECSSD \cite{shi2015hierarchical} \hspace{1mm}} \\ \cline{2-10}
      & { Acc } & {IoU} & {$maxF_{\beta}$} & { Acc } & {IoU} & {$maxF_{\beta}$} & {Acc } & {IoU} & {$maxF_{\beta}$} \\ \midrule
      TokenCut & .880 & .533 & .600 & .903 & .576 & .672 & .918 & .712 & .803 \\
      TokenCut + GF & .880 & .553 & .676 & .907 & .597 & .740 & .925 & .740 & .861 \\
      TokenCut + BF & \textbf{.897} & \textbf{.618} & \textbf{.697} & \textbf{.914} & \textbf{.624} & \textbf{.755} & \textbf{.934} & \textbf{.772} & \textbf{.874} \\
      \midrule
      MOVE & .923 & .615 & .712 & .950 & .713 & .815 & .954 & .830 & .916 \\
      MOVE + GF & .925 & .615 & .631 & \textbf{.952} & \textbf{.726} & .813 & \textbf{.958} & \textbf{.843} & \textbf{.923} \\
      MOVE + BF & \textbf{.931} & \textbf{.636} & \textbf{.734} & .951 & .687 & \textbf{.821} & .953 & .801 & .916 \\
    \bottomrule
  \end{tabular}
  \label{table:ablation_gf}
\end{table}

\subsection{Limitations}
\begin{figure}[t]
  \centering
  \includegraphics[width=\linewidth]{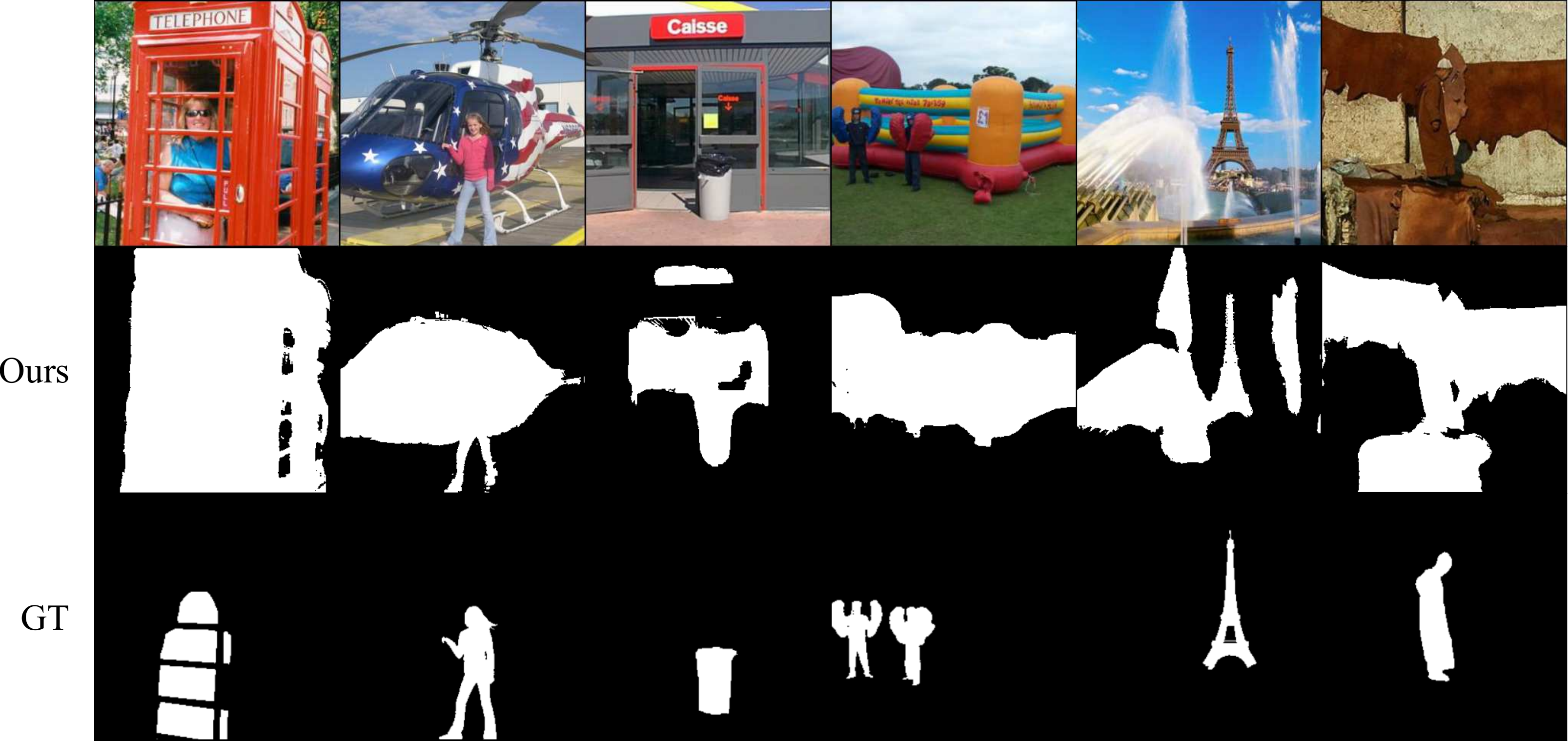}
  \caption{Some of the failure cases of our method from DUTS-TE \cite{wang2017learning}. In most cases the saliency is unambiguous or the background and the foreground are almost indistinguishable.}
  \label{fig:main_fail}
\end{figure}

Our method exhibits challenges in scenarios where saliency within an image is unambiguous or when multiple objects are present in the scene. 
Some visual examples can be seen in Fig. \ref{fig:main_fail}, where our method tends to segment all objects in the scene, whereas human annotators are inclined to select smaller and easily movable objects. Moreover, in instances where the salient object and the background share visual similarities our method encounters difficulty, making it challenging to distinguish between them accurately. 



\end{document}